\RequirePackage{fix-cm}
\documentclass[twocolumn]{svjour3}                     
\smartqed  
\usepackage{graphicx}
\usepackage{amsmath}
\usepackage[mathlines]{lineno}
\usepackage{physics}
\usepackage{mathtools}
\usepackage{booktabs}
\usepackage[misc]{ifsym}
\usepackage{csquotes}
\usepackage{listings}
\usepackage{makecell}
\usepackage{float}

\usepackage{pifont}
\newcommand{\cmark}{\ding{51}}     
\newcommand{\xmark}{\ding{55}}     

\usepackage{xcolor}

\definecolor{commentsColor}{rgb}{0.497495, 0.497587, 0.497464}
\definecolor{keywordsColor}{rgb}{0.000000, 0.000000, 0.635294}
\definecolor{stringColor}{rgb}{0.558215, 0.000000, 0.135316}
\definecolor{darkblue}{rgb}{0.0,0.0,0.6}

\definecolor{maroon}{rgb}{0.5,0,0}
\definecolor{darkgreen}{rgb}{0,0.5,0}
\lstdefinelanguage{XML}
{
  basicstyle=\ttfamily,
  morestring=[s]{"}{"},
  morecomment=[s]{?}{?},
  morecomment=[s]{!--}{--},
  commentstyle=\color{darkgreen},
  moredelim=[s][\color{black}]{>}{<},
  stringstyle=\color{blue},
  identifierstyle=\color{maroon},
  otherkeywords={Constants,SPH_Particle,SPH_Fluid,SPH_KernelGradCorrection_State,Interactions,SPH_Interaction_Neighborhood_Search,Scenarios,SPH_Scenario_Interaction_Fluid,Nested,SPH_Interaction_Fluid,SPH_Wendland_Kernel},
}
\journalname{Multibody System Dynamics}
\begin{document}

\title{Experiments with Large Language Models on Retrieval-Augmented Generation for Closed-Source Simulation Software}
\titlerunning{Experiments with LLMs on RAG for Closed-Source Simulation Software}

\author{
  Andreas Baumann (0000-0001-5831-4510) \and
  Peter Eberhard (0000-0003-1809-4407)
}
\authorrunning{Andreas Baumann \and Peter Eberhard}

\institute{
    \Letter\ Peter Eberhard\\
    \email{peter.eberhard@itm.uni-stuttgart.de}\\
    Andreas Baumann \and Peter Eberhard
    \at Institute of Engineering and Computational Mechanics, University of Stuttgart, Pfaffenwaldring 9, 70569 Stuttgart, Germany
}

\date{Received: date / Accepted: date}

\maketitle

\begin{abstract}
Large Language Models (LLMs) are tools that have become indispensable in development and programming. However, they suffer from hallucinations, especially when dealing with unknown knowledge. This is particularly the case when LLMs are to be used to support closed-source software applications. Retrieval-Augmented Generation (RAG) offers an approach to use additional knowledge alongside the pre-trained knowledge of the LLM to respond to user prompts. Possible tasks range from a smart-autocomplete, text extraction for question answering, model summarization, component explaining, compositional reasoning, to creation of simulation components and complete input models.
This work tests existing RAG systems for closed-source simulation frameworks, in our case the mesh-free simulation software Pasimodo. Since data protection and intellectual property rights are particularly important for problems solved with closed-source software, the tests focus on execution using local LLMs. In order to enable smaller institutions to use the systems, smaller language models will be tested first.
The systems show impressive results, but often fail due to insufficient information. Different approaches for improving response quality are tested. In particular, tailoring the information provided to the LLMs dependent to the prompts proves to be a significant improvement. This demonstrates the great potential and the further work needed to improve information retrieval for closed-source simulation models.

\end{abstract}

\section{Introduction}
\label{intro}
Large Language Models (LLMs) are becoming increasingly important and have become an integral part of everyday use. Writing texts or developing programming codes are just two examples that are generated using LLMs in the shortest time using questions and instructions written in natural human language. Among the most commonly used tools are ChatGPT or GitHub Copilot.

Overly simplified, however, it must be stressed that classic LLMs do not understand the posed tasks but rather predict the response on the prompt based on learned word and sentence sequence relationships. To obtain these relationships of natural language, large training sets are required. This allows LLMs to answer also prompts which were not included in its training set.

However, for knowledge-intensive tasks this often leads to wrong answers, so-called \emph{hallucinations} \cite{Lewis20,Ji2023}. Examples of knowledge-intensive tasks include question answering, code generation, and simulation model creation. The challenges in these tasks arise from knowledge that rarely appears or only once \cite{Kandpal23} or outdated knowledge created after cut-off of their training data \cite{Zhao}.
An example of automatically generating multibody simulation models based on problem descriptions in natural language using different LLMs is shown in \cite{Gerstmayr2024}. The experiments show that LLMs are able to sketch simulation models in Python using SciPy for basic dynamic problems. However, syntax and model errors often occur, which can be significantly reduced by adding suitable information to the context of the LLM. This was particularly evident for the open-source multibody simulation framework Exudyn \cite{Gerstmayr2023}.

Retrieval-Augmented Generation (RAG) \cite{Lewis20} provides an approach to address these challenges by combining additional external knowledge with the application of LLMs. Based on the input prompt, matching documents are retrieved e.g. from a precompiled database and appended to the input prompt before sending it to the LLM for response generation. As knowledge bases, a collection of documents, knowledge-graphs, or the internet can be used. RAG shows success in reducing the risk of hallucinations \cite{Shuster2021,Wu2024}, as relevant information for knowledge-intensive tasks can be provided.

RAG is applied in many fields from multi-language question answering \cite{Ranaldi25}, to medical question answering and summarization \cite{Miao2025}, and legal technologies e.g. for legal research \cite{Hindi2025}. In the following the focus is laid on machine learning approaches for simulation model creation using natural language.

An LLM system using RAG to create simulation models from natural language for the computational fluid dynamics software OpenFOAM is presented in \cite{Pandey2025}. Based on the user query describing the simulation setup, a RAG database for domain-specific OpenFOAM knowledge is consulted. The database is built from OpenFOAM tutorial case descriptions. Afterwards, an LLM is used to reason and generate a plan before the simulation model is set up and executed. Any error during the execution is appended to the original query and fed back.
The authors of \cite{Pandey2025} conclude that RAG is essential for model setup as otherwise the agent fails in most cases. The LLM is able to perform certain changes on the simulation models, such as changing boundary conditions or adjusting the mesh resolution. An application therefore is presented in transformation of existing simulation models to newer, yet incompatible, simulation software versions.
However, the authors argue that human oversight is crucial to validate the LLM's output for accuracy and correctness.

The RAG LLM system build for OpenFOAM is extended in \cite{FengEtAl25} to a multi-agent framework using LLMs designed to fully automate CFD simulations from natural language. A pre-processing agent interprets the users query and determines the simulation scope, i.e., a single evaluation or a multi-case parametric study, and selects the applied meshing strategy. Next, an LLM creates a detailed prompt needed to create the simulation model, which is done by the RAG LLM system presented in \cite{Pandey2025}. Any error on execution is fed back to the model creating LLM until the simulation succeeds. Afterwards, a post-processing agent is tasked to generate Python scripts for output data extraction and visualization. The authors report a high success rate and reproducibility rate across different simulations. The framework is able to perform automated parametric studies, e.g. varied physical properties like viscosity and contact angles.

For multibody systems (MBS), a RAG-like system is used in \cite{Moeltner2025} to automate the creation, evaluation and self-validation of mechanical simulation models using LLMs. The work focuses on MBS simulation models in Python using the Exudyn simulation code. Based on mechanical problems, like a flying mass point or a slider crank example, thousands of test templates are created by randomizing the problem parameters, such as mass or velocity. An LLM-agent proposes necessary simulation components retrieved from a predefined list of 22 available elements, based on the textual problem description, i.e. the input prompt. A tailored prompt is created consisting of textual problem description, code examples, and explanations of the previously selected elements to allow in-context learning of the LLM. The LLM is finally tasked to create the complete python simulation code. Executability and correctness are evaluated by automatically executing the created code and comparing the results with an expert-written ground truth.
This is further extended by testing the capabilities of LLMs to self-validate. Therefore, an LLM-agent chooses a set of evaluation methods from a predefined list, e.g. evaluation of the static equilibrium or the eigenfrequencies. An LLM creates the simulation code for the chosen evaluation method. The numerical results upon execution are extracted, simplified (resampling, and reduced resolution by rounding to 4 digits), and converted into textual description. An evaluator LLM receives the original problem description, the evaluation method chosen, and the simplified numerical results (not the code) for scoring. The framework shows a sufficiently high self-validation, but points out serious deficiencies in relation to geometric understanding of the problem. The LLM is incorrectly assuming common (human-chosen) simulation conventions (gravity in negative axis direction) and misinterpretats geometric descriptions, especially different coordinate systems and positions of joints.

The presented applications, although benefiting from RAG providing additional context, are developed for open-source frameworks which could already have been part of the training data used by commercial LLMs \cite{Gerstmayr2024}. The continued development of even larger LLMs by commercial companies, therefore, will be likely familiar with the concepts applied by the open-source frameworks. In contrast, closed-source simulation frameworks face significant challenges in that regard as the proprietary concepts, specific syntax, and interfaces are not part of the public training datasets. Therefore, any LLM response will be guessed based only on other publicly available software packages. This assumption will later be tested. 

In consequence, further research is required how closed-source frameworks can be used beneficially with LLMs to assist their users. Extending the abilities of LLMs to closed-source frameworks would be a tremendous value in automating simulation setup, explaining error messages, and many more. However, critical concerns arise from applying LLMs as the protection of intellectual property, data privacy, and independence from a few large model providers, which are just some reasons why these software packages or other knowledge remain internally. The same applies for other internal knowledge, ranging from software projects to internal project reports.

One factor that needs to be emphasized in further research is therefore, how LLMs can be used on local hardware. Alternatives for RAG to include syntax and knowledge about private-source frameworks would be the in-house fine-tuning of open-weight models \cite{Ali2024,llama_herd} or even complete training of own LLMs. However, this is infeasible due to limited available internal datasets and, even more, lack of the massive computational resources \cite{Llama3.2} required to do so for many companies and research groups. Thus, these approaches are not further considered in this work.

Instead, RAG promises a more viable and efficient path to leverage LLMs for internal closed-source simulation frameworks. RAG allows for the integration of proprietary knowledge at inference time, using local databases compiled from internal documentation, code examples, and internal reports. This approach not only addresses the LLM's knowledge gap but also allows the system to be operated locally. An increasing number of open-weight LLMs \cite{Ali2024,llama_herd,Gemma3} enable execution on local hardware. This ensures that sensitive intellectual property and other internal knowledge remain secure and are never disclosed to third-party model providers.

This work applies RAG on closed-source simulation software in order to assess its further application in leveraging LLMs for them. Different application cases with increasing complexity are considered to evaluate further requirements and state of current applicability: text extraction for question answering, structured text extraction, model summarization, component explaining, compositional reasoning, and finally creation of simulation components up to complete models. These are encountered in the preprocessing of simulations, but LLMs can be leveraged to create scripts for post-processing results.

The experiments will be carried out with different levels of documents provided to the RAG system. Initially, only existing documentation and examples available for the closed-source simulation software will be used. Based on this, different approaches to improve the quality of SPH simulations with LLMs are tested, such as adjusting the system prompt, providing additional more abstract knowledge, and improved augmentation with tailored information. The additional knowledge supplied consists of more abstract internal reports, in this case publications that have used the simulation software. These are supplied to the RAG system, and in a second step, general but domain specific information in the form of standard literature is added.

The simulation software Pasimodo \cite{Fleissner2010} is used in this work as an example for a closed-source simulation software. Pasimodo is a program package for particle-based simulation methods, such as Smoothed Particle Hydrodynamics (SPH) \cite{Violeau2016} and the Discrete Element Method (DEM) \cite{CundallStrack79}. The use of particle software also has the novelty that particle systems differ significantly in size from the presented literature, as multibody systems contain a small number of bodies, forces, and links, whereas particle systems consist easily of 100,000 particles. This pushes pure redundant definitions to their limits and more complex model definitions become necessary. The obtained results will be put into context to transfer the findings onto other closed-source software.

As RAG LLM systems, we use NotebookLM from Google \cite{notebooklm}, which utilizes the large-scale LLM Gemini 2.5 Flash, and the open-source solution AnythingLLM \cite{AnythingLLM} with different locally-executed open-weight LLMs. The focus in this work is laid on data protection, thus no source code is used. To pursue this and to allow other small organizations to assess applicability, only limited resources are assumed to be available for the locally hosted LLMs. Therefore, the small-scale LLMs, i.e. Llama~3.2 3B \cite{Llama3.2}, Gemma~3 4B, and Gemma~3 27B \cite{Gemma3}, are compared, which can be executed on consumer-grade GPUs.

This work is structured as follows. Section~\ref{sec:llm} provides an overview about LLMs in general and tests any prior knowledge of the tested LLMs regarding Pasimodo. The different steps of usage for the RAG LLM system are tested in Section~\ref{sec:rag} and in Section~\ref{sec:transfer} the results are generalized for application on other closed-source simulation software and for the further development of open-weight LLMs and RAG systems. A summary and conclusions for further work are discussed in Section 5.

\section{Application of Large Language Models}
\label{sec:llm}

Large Language Models are a type of machine learning model designed to generate text based on user input. The user input, also called prompt, can be natural language, which significantly simplifies the application. LLMs are trained on large sets of data, hence the name \emph{large}, to learn word and sentence sequence relationships. In oversimplified terms, an LLM learns to predict the most likely text sequence following the previous sequences. However, the LLM does not actually comprehend the problem behind the user input, it can detect relationships and patterns in text which allows it to create responses on user prompts which have not been part of its training set.

In the following, we give a brief introduction in the concept behind LLMs and the computational resources needed for them. Afterwards, we test their knowledge about a closed-source simulation software which has not been part of their training data.

\subsection{Brief Introduction into Large Language Models}

A complete introduction into LLMs and their architecture is beyond the scope of this publication. Therefore, a summary is provided of the main characteristics to understand the concept and the cornerstones of their application for RAG and closed-source software projects. Challenges for computers arise as natural language is diverse and complex as, e.g., the meaning of words change based on their context in a sentence.
To process the natural language input, called prompt, it is split into so-called tokens, words, word-parts, or characters, and converted into a vector representation, a so-called embedding, allowing its processing. Transformers \cite{AttentionIsAllYouNeed} introduced the attention mechanism to determine the importance each word in a sequence carries in relation to every other word of the same sequence. These relations are learned during the training phase requiring large training data to detect patterns, e.g. grammatical order of sentences. The attention mechanism is performed multiple-times (multi-head) each in parallel with different weight matrices where each of them can be trained to focus on different types of relationships. Additionally, a positional encoding is added to its initial embedding to account for word order in the input sequence.

Traditional transformers were originally built on two main components, an encoder and a decoder. The encoder creates a contextual representation of the input sequence by applying multiple layers of the multi-head self-attention mechanisms using the full input sequence and feedforward neural networks to process the detected relationships on each token's representation. The decoder creates the output sequence one token at a time. It also consists of multiple layers of multi-head self-attention mechanisms, a cross-attention mechanism attending to the encoder's output, and as before feedforward neural networks to process the output of the attention layers. In contrast to the encoder, the self-attention mechanism of the decoder can only access previous tokens in the output-sequence.

Based on traditional transformer architecture, significant progress developed by the introduction of encoder-only models, like BERT \cite{Devlin2019}, and decoder-only models, like Llama \cite{llama_herd}, where the respective other part of the traditional architecture is discarded. This simplifies the design and training of LLMs. Encoder-only models are trained on masking parts of the input sequence and its reconstruction, whereas decoder-only models are trained on predicting the next word. After this initial pretraining, an additional instruction tuning is performed to fine-tune the models to follow instructions and generate helpful responses.

The recent development of mixture of experts (MoE) \cite{Shazeer17} models are another class based on the transformer architecture. Instead of feedforward neural networks, certain layers of the architecture consists of a number of \emph{experts}, which each is a dense neural network \cite{sanseviero2023moe}. However, tokens are not sent to the full layer, but only to selected experts for processing. The decision which expert is selected is pretrained as the rest of the model is. As a lower number of parameters is required for processing a token, an MoE is pretrained and evaluated faster than a dense LLM. However, the required memory is higher when all expters are loaded into memory. Examples of MoE models are the Llama 4 models \cite{Llama4} or Mixtral 8x7B \cite{Mixtral23}.

Depending on the number of words known to the LLM and the number of layers, the size of current LLMs reaches into the billions of parameters. Usually the name indicates the model size, where the word billion is shortened to \emph{B} and corresponds to $10^9$. Table~\ref{tab:llm_overview} lists common LLMs, the numbers of parameters of their model, and the maximum input length. Gemini 2.5 Flash is an LLM offered by Google, whereas the Llama models provided by Meta or Gemma models also provided by Google are open-weight models, which can be evaluated locally. They are used later as foundation of the compared retrieval-augmented generation systems.

The training of an LLM requires massive computational resources and a sufficient amount of memory. Llama~3.2 3B required up to 460k GPU hours of training on data center Nvidia H100-80GB GPUs \cite{Llama3.2}. Such training expenses are not feasible for individual research groups.
After the LLM has been trained, the required amount of memory to operate the LLM reduces to storing the model parameters, either in RAM or in the VRAM on a GPU and performing its evaluation. The required memory can be further reduced by applying a method called quantization. Therefore, the parameters are converted from high precision 32-bit or 16-bit floating point numbers to lower-precision format \cite{Liu24}, like 8-bit or 4-bit integers. The 3-billion-parameter model Llama~3.2 3B for example, takes about 1.8 GB when the parameters are stored in 4-byte integer format. However, this comes with the price of less accuracy. This allows to fit small models, like Llama~3.2 3B or Gemma~3 27B, on current consumer-grade GPUs and their evaluation can be done in reasonable times. Larger models with 70B parameters or more require more specialized GPUs, though.

\begin{table}[ht!]
  \center
  \caption{Overview of the LLMs used in the first example.}
  \label{tab:llm_overview}
  \begin{tabular}{p{2.35cm}p{2.1cm}p{2.7cm}}
    \toprule
    LLM & parameters & max. input/context length \\
    \midrule
      ChatGPT 4o & 1 - 1.8 trillion\footnotemark & 128,000 tokens \cite{OpenAi4o} \\
      Gemini 2.5 Flash & ?\footnotemark & 1,048,576 tokens \cite{GeminiAPIDocs2.5} \\
      Llama~3.2 3B \newline (local) & 3.21 billion \cite{Llama3.2} & 128,000 tokens \cite{Llama3.2} \\
      Gemma~3 4B \newline (local) & 4.3 billion \cite{Gemma3} & 128,000 tokens \cite{Gemma3ModelCard} \\
      Gemma~3 27B \newline (local) & 27.43 billion \cite{Gemma3} & 128,000 tokens \cite{Gemma3ModelCard} \\
    \bottomrule
  \end{tabular}
\end{table}
\footnotetext[1]{Officially OpenAI does not disclose the number of parameters of ChatGPT, thus only estimations are available \cite{ExploringTopics}.}
\footnotetext[2]{Google does not disclose the parameter size.}

\subsection{Example 1: Knowledge about Pasimodo}

Depending on their training data, LLMs are able to answer questions reliable without further data. Therefore, the LLMs listed in Table~\ref{tab:llm_overview} are asked about the closed-source simulation software Pasimodo which is used as an example and their responses are listed in Table~\ref{tab:example_pasi}.

Whereas the local evaluation of the Llama~3.2 3B model diverts and offers a summary on the fictional character Quasimodo, the two large-scale hosted LLMs ChatGPT 4o and Gemini 2.5 Flash performed a web search about Pasimodo. Both compiled their answers from the websites available on Pasimodo, which they listed as sources, too. Both Gemma~3 models made up their answer and provide weblinks for additional information, which do not exist.

Apparently, all LLMs did not include any information about Pasimodo in their training. The web searches of ChatGPT and Gemini indicate that both are not pure LLMs but rather have already capabilities to retrieve additional information before generating their answers. However, as the public information on Pasimodo is limited, more detailed questions were not asked. The next section will present an approach to leverage more internal knowledge about Pasimodo for answering questions and tasks using LLMs.

\begin{table*}[ht!]
  \center
  \caption{Answers on the question: \enquote{What do you know about Pasimodo?}.}
  \label{tab:example_pasi}
  \begin{tabular}{p{3cm}p{13cm}}
    \toprule
    LLM & output (shortened)\\
    \midrule
    Llama~3.2 3B & \enquote{I think you meant "Quasimodo". Quasimodo is the fictional protagonist of Victor Hugo's novel "The Hunchback of Notre-Dame", published in 1831. He is a deformed and lonely bell-ringer who lives in the famous Notre-Dame Cathedral in Paris. [\dots]} \\
    Gemma~3 4B & \enquote{Pasimodo is a fascinating and somewhat unique project originating from Russia, centered around creating incredibly detailed, miniature dioramas depicting scenes from popular video games, movies, and anime.} [\dots] \\
    Gemma~3 27B & [\dots] \enquote{Pasimodo is an AI-powered platform that specializes in creating high-quality, consistently-styled 3D animations and videos from text prompts. It's unique because it's *not* about generating individual images or short clips, but rather creating *long-form, coherent, and cinematic* animations. Think of it as an AI director and animator rolled into one.} [\dots] \\
    ChatGPT 4o & \enquote{Searched 3 sites: Pasimodo is a software package designed for particle-based simulation methods, particularly suited for the study of granular media and Smoothed Particle Hydrodynamics (SPH) simulations. Developed by the Institute of Engineering and Computational Mechanics (ITM) at the University of Stuttgart, [\dots]} \\
    Gemini 2.5 Flash & \enquote{Pasimodo is a program package for particle-based simulation methods, primarily developed at the Institute of Engineering and Computational Mechanics (ITM) of the University of Stuttgart, Germany. [\dots]} various ITM webpages related to Pasimodo are listed as sources\\
    \bottomrule
  \end{tabular}
\end{table*}

\section{Application of Retrieval-Augmented Generation}
\label{sec:rag}

Retrieval-Augmented Generation (RAG) \cite{Lewis20} refers to an approach where additional information is retrieved and provided additionally to the LLM. In consequence, more knowledge can be used than the internal memory of the LLM obtained through pretraining. Next follows a brief introduction into RAG, which is followed by testing available out-of-the-box RAG solutions for leveraging LLMs for internal closed-source simulation frameworks.

\subsection{Brief Introduction into Retrieval-Augmented Generation}
A RAG system combines the pretrained abilities of an LLM with an additional knowledge database to provide up-to-date and context-specific information. It combines the information retrieval with natural language processing. The additional information can be internal documents, wiki or internet pages, or existing databases. The user prompt is used to perform a semantic search on the document database, in some cases also a fuzzy search or a search engine in the case of webpages can be used.

\begin{figure}[htb]
  \centering
  \def\svgwidth{.5\textwidth}

  \begingroup%
    \makeatletter%
    \providecommand\color[2][]{%
      \errmessage{(Inkscape) Color is used for the text in Inkscape, but the package 'color.sty' is not loaded}%
      \renewcommand\color[2][]{}%
    }%
    \providecommand\transparent[1]{%
      \errmessage{(Inkscape) Transparency is used (non-zero) for the text in Inkscape, but the package 'transparent.sty' is not loaded}%
      \renewcommand\transparent[1]{}%
    }%
    \providecommand\rotatebox[2]{#2}%
    \newcommand*\fsize{\dimexpr\f@size pt\relax}%
    \newcommand*\lineheight[1]{\fontsize{\fsize}{#1\fsize}\selectfont}%
    \ifx\svgwidth\undefined%
      \setlength{\unitlength}{120.11774826bp}%
      \ifx\svgscale\undefined%
        \relax%
      \else%
        \setlength{\unitlength}{\unitlength * \real{\svgscale}}%
      \fi%
    \else%
      \setlength{\unitlength}{\svgwidth}%
    \fi%
    \global\let\svgwidth\undefined%
    \global\let\svgscale\undefined%
    \makeatother%
    \begin{picture}(1,0.38938042)%
      \lineheight{1}%
      \setlength\tabcolsep{0pt}%
      \put(0,0){\includegraphics[width=\unitlength,page=1]{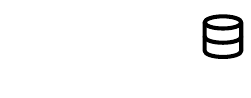}}%
      \put(0.46089152,0.31){\makebox(0,0)[lt]{\lineheight{1.25}\smash{\begin{tabular}[t]{l}[0.15,0.48,...,1.86]\end{tabular}}}}%
      \put(0.46089152,0.26267975){\makebox(0,0)[lt]{\lineheight{1.25}\smash{\begin{tabular}[t]{l}[0.34,0.23,...,2.24]\end{tabular}}}}%
      \put(0.46089152,0.175){\makebox(0,0)[lt]{\lineheight{1.25}\smash{\begin{tabular}[t]{l}[0.87,0.89,...,1.79]\end{tabular}}}}%
      \put(0,0){\includegraphics[width=\unitlength,page=2]{setup.pdf}}%
      \put(0.01667739,0.05408442){\makebox(0,0)[lt]{\lineheight{1.25}\smash{\begin{tabular}[t]{l}documents\end{tabular}}}}%
      \put(0.25358242,0.05408442){\makebox(0,0)[lt]{\lineheight{1.25}\smash{\begin{tabular}[t]{l}chunks\end{tabular}}}}%
      \put(0.51741103,0.05408442){\makebox(0,0)[lt]{\lineheight{1.25}\smash{\begin{tabular}[t]{l}embeddings\end{tabular}}}}%
      \put(0.84412272,0.0931371){\makebox(0,0)[lt]{\lineheight{1.25}\smash{\begin{tabular}[t]{l}vector\end{tabular}}}}%
      \put(0.82644307,0.05408241){\makebox(0,0)[lt]{\lineheight{1.25}\smash{\begin{tabular}[t]{l}database\end{tabular}}}}%
    \end{picture}%
  \endgroup%

  \caption{Creation of a database for RAG: documents are split into smaller chunks which are embedded into vector representation before adding them to the database.}
  \label{fig:rag_setup}
\end{figure}

Before using RAG with documents, the necessary database must be created. Using a database speeds up the retrieval of information relevant to the user prompt. A vector database is suitable for this purpose as it enables the similarity search upon retrieval with the embedded user prompt to find similar topics and thereby matching information. Depending on the data format of the documents, they are converted into plain text files first before embedding them into a vector representation. LLMs are limited in the input length they can handle, thus the input documents are also split into smaller parts, so-called chunks \cite{Gao2023}. Figure~\ref{fig:rag_setup} illustrates the creation of the neccessary database.

\begin{figure}[htb]
  \centering
  \def\svgwidth{.5\textwidth}

  \begingroup%
    \makeatletter%
    \providecommand\color[2][]{%
      \errmessage{(Inkscape) Color is used for the text in Inkscape, but the package 'color.sty' is not loaded}%
      \renewcommand\color[2][]{}%
    }%
    \providecommand\transparent[1]{%
      \errmessage{(Inkscape) Transparency is used (non-zero) for the text in Inkscape, but the package 'transparent.sty' is not loaded}%
      \renewcommand\transparent[1]{}%
    }%
    \providecommand\rotatebox[2]{#2}%
    \newcommand*\fsize{\dimexpr\f@size pt\relax}%
    \newcommand*\lineheight[1]{\fontsize{\fsize}{#1\fsize}\selectfont}%
    \ifx\svgwidth\undefined%
      \setlength{\unitlength}{120.11774826bp}%
      \ifx\svgscale\undefined%
        \relax%
      \else%
        \setlength{\unitlength}{\unitlength * \real{\svgscale}}%
      \fi%
    \else%
      \setlength{\unitlength}{\svgwidth}%
    \fi%
    \global\let\svgwidth\undefined%
    \global\let\svgscale\undefined%
    \makeatother%
    \begin{picture}(1,0.46051686)%
      \lineheight{1}%
      \setlength\tabcolsep{0pt}%
      \put(0,0){\includegraphics[width=\unitlength,page=1]{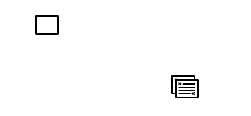}}%
      \put(0.20680958,0.3353272){\makebox(0,0)[lt]{\lineheight{1.25}\smash{\begin{tabular}[t]{l}?\end{tabular}}}}%
      \put(0,0){\includegraphics[width=\unitlength,page=2]{retrieval.pdf}}%
      \put(0.13294455,0.28496412){\makebox(0,0)[lt]{\lineheight{1.25}\smash{\begin{tabular}[t]{l}prompt\end{tabular}}}}%
      \put(0.36379303,0.28121154){\makebox(0,0)[lt]{\lineheight{1.25}\smash{\begin{tabular}[t]{l}embedding\end{tabular}}}}%
      \put(0.40786228,0.01873161){\makebox(0,0)[lt]{\lineheight{1.25}\smash{\begin{tabular}[t]{l}LLM\end{tabular}}}}%
      \put(0.14723053,0.01873161){\makebox(0,0)[lt]{\lineheight{1.25}\smash{\begin{tabular}[t]{l}reply\end{tabular}}}}%
      \put(0.61435341,0.23981468){\makebox(0,0)[lt]{\lineheight{1.25}\smash{\begin{tabular}[t]{l}similarity search\end{tabular}}}}%
      \put(0.61989173,0.01924361){\makebox(0,0)[lt]{\lineheight{1.25}\smash{\begin{tabular}[t]{l}retrieved chunks\end{tabular}}}}%
      \put(0.30840364,0.35124907){\makebox(0,0)[lt]{\lineheight{1.25}\smash{\begin{tabular}[t]{l}[0.78,0.90,...,1.80]\end{tabular}}}}%
      \put(0,0){\includegraphics[width=\unitlength,page=3]{retrieval.pdf}}%
    \end{picture}%
  \endgroup%

  \caption{Retrieval-augmented generation: based on the embedded prompt, similar chunks are search within the vector database, which are added to the prompt and sent to the LLM for reply generation.}
  \label{fig:rag_retrieval}
\end{figure}

For new user prompts, the retriever searches based on the embedded user prompt for close matches within the vector database. The retrieved documents with the highest similarity to the prompt are combined with the original prompt, which is called augmentation, before they are handed to the LLM for the response generation \cite{Gao2023}. For the response generation, any existing LLM can be used. As with the classic use of the LLM, previous dialog can be added as chat history too. Figure~\ref{fig:rag_retrieval} illustrates the retrieval and augmentation based on the user prompt.

Besides the naive RAG approach, more advanced techniques exists as challenges with selection of misaligned or irrelevant chunks exist. These approaches e.g. include an additional re-ranking of the retrieved information to improve the relevance of the retrieved information \cite{Gao2023}.

\subsection{System Setup}
For the experiments, the out-of-the-box solutions NotebookLM \cite{notebooklm} (version November 2025), a RAG tool from Google using Gemini 2.5 Flash, and AnythingLLM \cite{AnythingLLM}, an open-source RAG application, which can be run locally using a local LLM instance, are used. NotebookLM does not provide much setup options allowing only the selection of the uploaded sources which should be used for answering the posted questions. AnythingLLM allows for more modifications such as choosing the used LLM, the vector database storing the provided data, and more. As the focus is set on evaluating the available out-of-the-box performance of these systems and not developing a new RAG technique, only minimal changes besides connecting local LLMs are applied. The maximum token limit is increase to 8192, which is also set for Ollama. To enable re-ranking to retrieve better results, the accuracy optimized search preference is selected. Further, the temperature parameter of the LLMs is set to zero, which should cause the LLM to take the token with the highest probability \cite{PeeperkornEtAl24}.

As local LLMs, the small-scale open-weight LLMs Llama~3.2 3B \cite{Llama3.2}, Gemma~3 4B \cite{Gemma3}, and Gemma~3 27B \cite{Gemma3} are used. As the focus is laid on data protection for small organizations, the selected LLMs are chosen to allow execution on consumer-grade GPUs. In contrast, NotebookLM does provide the assessment of RAG capabilities on high-performance setups. Table~\ref{tab:system_properties} summarizes the local hardware setups applied.

\begin{table}[ht!]
  \center
  \caption{Local setups for executing AnythingLLM.}
  \label{tab:system_properties}
  \begin{tabular}{p{.75cm}p{2.9cm}p{2.6cm}}
    \toprule
    & setup 1 & setup 2\\
    \midrule
      CPU & \raggedright AMD Ryzen Threadripper 3970X & Intel i7-13700K\\
      RAM & 256 GB & 128 GB\\
      GPU & \raggedright NVIDIA GeForce RTX 4070 (12 GB) & NVIDIA GeForce RTX 4090 (24 GB)\\
      \addlinespace
      \multicolumn{3}{l}{LLM Provider: Ollama v0.12.6 \cite{ollama}} \\
      \multicolumn{3}{l}{RAG Software: AnythingLLM v1.8.5 \cite{AnythingLLM}} \\
    \bottomrule
  \end{tabular}
\end{table}

\subsection{Data Preprocessing}
The experiments are conducted using a collection of input examples of Pasimodo and a documentation on the most important points written as wiki including multiple tutorials for creating models with Pasimodo. The same data is supplied to new students as introduction and first steps to get familiar with Pasimodo. Additionally, the API reference for model input files, which is accessible to students as searchable web catalog, is included in the collection as formatted summary.

The data needs to be pre-processed before adding it to the RAG systems. NotebookLM has a limit of 50 sources, e.g. pdf, text, or markdown-files, which each can contain up to 500,000 words. First, the repositories containing the input examples and the wiki were converted into a single text file. This text file was split into multiple files each containing 400,000 words, which were identified by being separated by spaces, equals signs, and slashes. In a first test, NotebookLM removed all XML markup from the uploaded plain text files. This was problematic as the input files for Pasimodo are written in XML, therefore, causing a corrupted database. To prevent the XML markup removal, the plain text files were printed as pdf files before uploading them to NotebookLM. The same pdf files were added to AnythingLLM without further processing.

\subsection{Test Structuring}
The test scenarios are based on the areas of application of LLMs for closed-source simulation software, starting from question answering and smart autocomplete to complete model creation. They are structured in text extraction for question answering, structured text extraction, model summarization, component explaining, compositional reasoning, and finally creation of simulation components and complete models.

Now, the first example, Table~\ref{tab:example_pasi}, is repeated using the RAG systems to verify that the provided information is available for answer generation. Then, exemplary beginner questions are asked requiring text extraction from the provided documentation, such as those that might arise for a new user. This not only supports beginners but also reduces the load on experts, as first-level support is provided by the RAG LLM system. The questions are centered on what is a particle, what particle is used to model a fluid or solid body, how multiple particles are generated, and how an inflow is defined.

The second set of example prompts require the extraction of structured text, e.g. the structure of XML model components: how is an SPH particle or the SPH interaction defined. This is followed by tasks to explain provided excerpts from simulation input files. Afterwards, the RAG LLM system is tasked to summarize complete input files of simulation models starting from an example taken from the initial collection of input examples used to compile the database. The models increase in complexity and include different areas where Pasimodo is applied ranging from DEM to SPH with heat conduction, a coupling test using the Functional Mockup Interface (FMI) before inserting a multi-file simulation model of a dam break example \cite{GomezGesteira2010}. As NotebookLM does not support attaching additional files to the prompt, the simulation models are added to the prompt. However, the maximum prompt length for NotebookLM prevents testing the larger simulation models. AnythingLLM with the open-weight LLMs, in contrast, is capable of processing the larger models. Here, the maximum context length of Ollama and the respective LLMs set the limit.

The questions are deliberately vague, as it takes expertise to formulate specific, targeted prompts which we do not assume for beginners. The final example to create a complete simulation model is formulated in detail to give clear instructions to the LLM. Table~\ref{tab:prompts} lists the complete set of prompts.

\subsection{Results for Software-specific Documents}
Table~\ref{tab:results_base} summarizes the results for the different LLMs and RAG systems tested when providing a collection of input examples of Pasimodo, the internal wiki, and the API reference as knowledge sources. The repeated Example~1 shows that, the systems are able to retrieve the provided information and answer the question about Pasimodo with more details. NotebookLM and AnythingLLM also provide citations, what parts of the sources were retrieved for generating the response.

\begin{table}[H]
  \center
  \caption{Prompts selected for testing the different areas of application of LLMs for closed-source simulation software.}
  \label{tab:prompts}
  \begin{tabular}{lp{7.25cm}}
    \toprule
    \# & Prompt \\
    \midrule
      \multicolumn{2}{l}{\textbf{text extraction for question answering}} \\
      1.1 & What do you know about Pasimodo? \\
      1.2 & What is a particle? \\
      1.3 & What particle should I use to model a fluid? \\
      1.4 & What particle should I use to model a solid body? \\
      1.5 & What component can be used to generate multiple particles? \\
      1.6 & How do I define an inflow? \\

      \addlinespace
      \multicolumn{2}{l}{\textbf{structured text extraction}} \\
      2.1 & How do I define an SPH particle? \\
      2.2 & How do I define the SPH interaction? \\
      2.3 & What integrator do I need for SPH? \\
      2.4 & How can I import CAD geometries into the model? \\
      2.5 & How do I define the properties of a liquid as oil? \\
      2.6 & How do I define the material as metal? \\

      \addlinespace
      \multicolumn{2}{l}{\textbf{component explaining}} \\
      3.1 & Explain the following code (Listing~\ref{lst:prompt31_code})\\
      3.2 & Explain the following code (Listing~\ref{lst:prompt32_code})\\
      3.3 & Explain the following code (Listing~\ref{lst:prompt33_code})\\
      3.4 & Explain the following code (Listing~\ref{lst:prompt34_code})\\

      \addlinespace
      \multicolumn{2}{l}{\textbf{model summarization}} \\
      4.1 & Summarize the following model Two Spheres DEM\\
      4.2 & Summarize the following model Three Spheres DEM\\
      4.3 & Summarize the following model SPH Heat Conduction Test\\
      4.4 & Summarize the following model FMI Force Distribution Test\\
      4.5 & Summarize the following model Dam Break Scenario\\

      \addlinespace
      \multicolumn{2}{l}{\textbf{compositional reasoning}} \\
      5.1 & How can I fix a particle in y and z direction? \\
      5.2 & How can I export the forces acting on the SPH particles using an HDF5 part output? \\
      5.3 & How to choose appropriate values for the sound velocity? \\
      5.4 & How can I create a cube of SPH particles with an edge length of 0.01 and a particle size of 0.001? \\
      5.5 & How can I define a rigid plate with edge length of 0.01, which is oscillating with a fixed amplitude of 0.001? \\

      \addlinespace
      \multicolumn{2}{l}{\textbf{creation of components up to complete models}} \\
      6.1 & Create a minimal working example which uses the component Influx\_External.\\
      6.2 & Create a complete input file for a 3D simulation of an oil droplet under gravity for one second. Use the plugins pplug\_SPH and pplug\_h5part. The droplet is spherical, has a diameter of 0.01, and consists of 10 particles in diameter. The gravity acts in negative z direction with a value of 9.81. Use the component SPH\_Particle to model the fluid which has a density of 845, a dynamic viscosity of 8.45e-3. Set the smoothing length of the SPH particle as 1.5 times the initial distance of the particles and the artificial sound velocity as 100. Assign the particle the mass of a cube with edge length of the initial distance. Use the SPH\_Extension\_Scenario and the SPH\_Scenario\_Interaction\_Fluid with SPH\_Interaction\_Fluid to add the fluid interactions between the SPH particles. Include the SPH Leapfrog integrator for time integration and create an h5part output which exports the position and velocity of the SPH particles at an interval of 1/10~s. Set up the simulation to run in parallel.\\
    \bottomrule
  \end{tabular}
\end{table}

By default, AnythingLLM is limited to retrieving 4 sources. NotebookLM cited a substantially greater number of sources, typically exceeding ten and, in extreme cases, reaching up to 197 sources.

In general, NotebookLM showed an impressive performance by correctly distinguishing particle types, detecting ambiguities, and proactively suggesting required plugins. In contrast, the answers from the three local LLMs differed in their quality, although the sources retrieved by AnythingLLM were the same for the LLMs. Especially Llama~3.2 3B went repeatably off-topic, e.g. included \emph{elementary particles} (electrons, protons, ...) in its explanation of particles (1.2) or hallucinated about macro-particles (1.3, 1.4), whereas Gemma~3 4B and mainly Gemma~3 27B focused instead on the retrieved knowledge. However, it also showed that the small number of sources taken into account led to hallucinations. For example, one source retrieved for prompt 1.4 contained a reference to \emph{ISPH\_Solid\_Particle}, which was interpreted by the LLMs e.g. as \emph{Implicit SPH} (Gemma~3 4B) or \emph{Iterated SPH} (Gemma~3 27B) instead of its actual meaning \emph{incompressible SPH}. Although no specifics on NotebookLM's architecture is available, we assume, that NotebookLM, in contrast, repeatably retrieves additional sources before generating the final response.

Furthermore, all local LLMs struggled significantly with technical accuracy and hallucinations due to incomplete information. For example, AnythingLLM retrieved for prompt 2.4 only partial information on importing CAD geometries as the corresponding wiki section was split into multiple chunks. Especially the small LLMs exhibited a high tendency to "fill in the blanks" with plausible-sounding but technically incorrect parameters and element names. Another problem became apparent where outdated interactions were retrieved during prompts and incorporated into the response, particularly with NotebookLM as it incorporated more sources.

In general, all LLMs, but Llama~3.2, performed acceptable on summarizing the provided code examples (prompt 3.1 - 3.4) although the retrieved references by AnythingLLM only contained chunks from input examples and no API reference. Therefore, the LLMs summarized the code by interpretation of the element and attribute names. In consequence, the larger LLM, Gemma~3 27B, was able to detect the auto-dissection of the rectangle or VTU as visualization toolkit unstructured grid format. Llama~3.2 had severe problems to distinguish between current input and the chat history, which is also part of the context sent to the LLM. Llama~3.2 stated that the input model of prompt 4.3, an SPH heat conduction test, contained a Hertzian contact interaction, which was actually part of the previous input model of prompt 4.2.

For the summary of the larger input files, NotebookLM failed as its maximum input prompt length prevented its application, whereas on the smaller input files, NotebookLM excelled. As AnythingLLM does not limit the input prompt length, the local LLMs are only limited by their maximum context length restricting the complete request (system prompt, user prompt, retrieved sources, and chat history). In general, the provided summaries are solid with minor errors from the smaller LLMs. For example, Llama~3.2 stated that the simulation model from prompt 4.1 exports the accelerations of the particles instead of the forces acting on them, besides their position, velocity, and mass. All local LLMs have issues with the two spheres example, where one is fixed, the other falls under gravity before it bounces off the first sphere. The LLMs either state that both spheres are fixed or miss the fixed position of the first one completely. Furthermore, the setting of gravity as zero vector confused Llama~3.2 and Gemma~3 3B into thinking gravity is applied in the simulation model, whereas Gemma~3 27B and NotebookLM identified it correctly as zero-gravity.

The local LLMs have performed adequately so far, but they reached their limits when attempting to combine independent knowledge. For example in reply to prompt 5.1, they failed at bitwise addition to lock axes (Llama~3.2) or contradicted themselves in their explanations (Gemma~3 4B). However, it became apparent in the course of further testing that missing references, such as the syntax of the input data for Pasimodo, caused the local LLMs to hallucinate model components or instructions to set up the model. NotebookLM, in contrast, was even able to extract correct approaches to choose the sound velocity (prompt 5.3), although detailed knowledge about its choice is missing in the documentation.

Surprisingly, all systems failed to create a minimal input example for the \emph{Influx\_External} component (prompt 6.1). Instead, the LLMs diverted to the similar component \emph{Inflow\_External} as, e.g., AnythingLLM retrieved the wrong information. An explanation could be the close similarity between the component names and the more frequent occurrence of \emph{Inflow\_External} in the provided collection of input examples, whereas \emph{Influx\_External} is only mentioned in the API reference.

The local systems, furthermore, failed on creating a complete simulation model due to hallucinations on the model syntax. The best simulation model was created by NotebookLM, but it is not executable because it had missing parameters in the definition of a component and a faulty SPH interaction. The model contained both an outdated SPH interaction and a correct definition, but the neighborhood search was missing. This is surprising because it previously summarized the SPH interaction completely in prompt 2.2 and also pointed out the necessary neighborhood search.

\begin{table}[ht!]
  \center
  \caption{Test evaluation using different local LLMs with AnythingLLM and NotebookLM. Qualitative results based on expert analysis weather the LLM output is correct  or contains errors.}
  \label{tab:results_base}
  \begin{tabular}{p{.5cm}p{1.25cm}p{1.25cm}p{1.25cm}c}
    \toprule
    \# & Llama~3.2 3B & Gemma~3 4B & Gemma~3 27B & NotebookLM \\
    \midrule
      \multicolumn{5}{l}{\textbf{text extraction for question answering}} \\
      1.1     & \cmark & \cmark         & \cmark          & \cmark \\
      1.2     & \xmark & \xmark         & \cmark          & \cmark \\
      1.3     & \xmark & \cmark         & \cmark          & \cmark \\
      1.4     & \xmark & \xmark         & \cmark / \xmark & \cmark \\
      1.5     & \cmark & \cmark         & \cmark          & \cmark \\
      1.6     & \cmark & \cmark/\xmark  & \cmark          & \cmark \\
      score   & 3/6    & 4/6            & 6/6             & 6/6 \\

      \addlinespace
      \multicolumn{5}{l}{\textbf{structured text extraction}} \\
      2.1     & \cmark  & \cmark           & \cmark          & \cmark \\
      2.2     & \xmark  & \xmark           & \cmark          & \cmark \\
      2.3     & \cmark  & \xmark           & \cmark          & \cmark \\
      2.4     & \xmark  & \xmark           & \xmark          & \cmark \\
      2.5     & \xmark  & \cmark / \xmark  & \cmark / \xmark & \cmark / \xmark \\
      2.6     & \xmark  & \cmark / \xmark  & \xmark          & \cmark \\
      score   & 2/6     & 3/6              & 4/6             & 6/6 \\

      \addlinespace
      \multicolumn{5}{l}{\textbf{component explaining}} \\
      3.1     & \xmark & \xmark           & \cmark & \cmark \\
      3.2     & \xmark & \cmark / \xmark  & \cmark & \cmark \\
      3.3     & \cmark & \cmark           & \cmark & \cmark \\
      3.4     & \cmark & \cmark           & \cmark & \cmark \\
      score   & 2/4    & 3/4              & 4/4    & 4/4 \\

      \addlinespace
      \multicolumn{5}{l}{\textbf{model summarization}} \\
      4.1     & \xmark        & \xmark          & \xmark & \cmark \\
      4.2     & \xmark        & \cmark          & \cmark & \cmark \\
      4.3     & \xmark        & \cmark          & \cmark & \xmark \\
      4.4     & \xmark        & \cmark / \xmark & \cmark & \xmark \\
      4.5     & \cmark        & \cmark          & \xmark & \xmark \\
      score   & 1/5           & 4/5             & 3/5    & 2/5 \\

      \addlinespace
      \multicolumn{5}{l}{\textbf{compositional reasoning}} \\
      5.1     & \xmark & \xmark & \cmark & \cmark \\
      5.2     & \cmark & \xmark & \cmark & \cmark \\
      5.3     & \xmark & \xmark & \xmark & \cmark \\
      5.4     & \xmark & \xmark & \xmark & \cmark \\
      5.5     & \xmark & \xmark & \xmark & \cmark \\
      score   & 1/5    & 0/5    & 2/5    & 6/6 \\

      \addlinespace
      \multicolumn{5}{l}{\textbf{creation of components up to complete models}} \\
      6.1     & \xmark & \xmark & \xmark & \xmark \\
      6.2     & \xmark & \xmark & \xmark & \xmark \\
      score   & 0/2    & 0/2    & 0/0    & 0/2 \\
    \bottomrule
  \end{tabular}
\end{table}

\subsection{Adjustments to Improve Result Quality}

The previous test results revealed that the LLMs are restricted by the provided specific information on the closed-source simulation software Pasimodo. Missing detailed information led to increased hallucinations, especially by the smaller LLMs, trying to "fill in the blanks". NotebookLM, in contrast, retrieved significantly more sources from the provided documentation compared to the retrieval of AnythingLLM, leading to significantly better results. The mid-scale LLM Gemma~3 27B, furthermore, exhibited a better understanding of the retrieved sources, which were the same for all LLMs using AnythingLLM, compared to the small LLMs.

In the following, different approaches are tested to improve the test result quality. The approaches instruct the LLM to focus on the simulation framework under consideration, provide additional information for retrieval or by tailoring the actual retrieval, and guide the LLM to solve the different fundamental problems in the results of previous tests. As NotebookLM provided excellent results for the tested prompts but has only limited options for customization, the approaches are restricted to the local setup using AnythingLLM and the local LLMs.

\subsubsection{System Prompt Tailoring}

For the ambiguous questions for text extraction, the local LLMs diverted, especially the smaller ones, and provided general responses, e.g., related to particles in prompt 1.2. Therefore, the system prompt of AnythingLLM, which is added to every user prompt, is specified that the LLM should focus on the \emph{mesh-free simulation software Pasimodo}, that its \emph{knowledge is limited to the terms of Pasimodo provided in the following section}, and it shall state if it is unsure and needs more information.

In general, the LLMs are less diverting and tend to less speculate based on element and attribute names compared to the base run, e.g. prompt 1.2 (no elementary particles) and prompt 1.4 (no hallucinations on ISPH). Llama~3.2 states repeatedly that it does not know before replying in a shorter manner than before. The responses of the LLMs on summarizing the models are a bit more focused, but not necessarily improved. However, the adjusted system prompt does not prevent hallucinations, which still occur, especially for compositional reasoning.

\subsubsection{Additional Non-Software-specific Documents}

The initial tests also revealed that essential knowledge about certain simulation settings is missing in the provided documentation, e.g. the choice of appropriate values for the sound velocity. The naive approach would be to extend the documentation to cover every parameter. However, this contradicts common practice, where software documentation, especially internal documentation, is incomplete because the focus is usually on development rather than customer support.

Instead, adding of additional knowledge, which is not directly software-specific, is tested. A significant amount of knowledge in using a software and the applications are usually written in (internal) project reports or, in the case of academia, in publications. Therefore, publications that have used the simulation software, are supplied to the RAG system. General but domain specific information in the form of standard literature will be added in a second step. The supplied literature of both steps is listed in Table~\ref{tab:additional_knowledge}.

\begin{table}[ht!]
  \center
  \caption{Additional publications supplied to the system. The (internal) project reports, in our case publications of the ITM, were done using Pasimodo. The general but domain specific information consists of standard SPH literature.}
  \label{tab:additional_knowledge}
  \begin{tabular}{p{1cm}p{6.5cm}}
    \toprule
    citation & description \\
    \midrule
    \multicolumn{2}{l}{\textbf{(internal) project reports}} \\
    \cite{Beck2013} & Modeling abrasive wear using SPH and DEM \\
    \cite{Eberhard2014} & Particle methods for description of solids and fluids \\
    \cite{Spreng2014} & Local adaptive discretization for SPH \\
    \cite{Schnabel17} & SPH for deep-hole drilling \\
    \cite{Eberhard2018} & SPH for ductile solid continua \\
    \cite{Gnanasambandham19} & Particle dampers using mesh-free simulation methods \\
    \cite{Shishova2019} & material orientation for solid continua description using SPH \\
    \cite{Sollich2022} & Recoil pressure boundary condition for laser beam welding \\
    \cite{Sollich2023} & Incompressible and weakly-compressible SPH for laser beam welding \\
    \cite{Sollich2023b} & Modeling of laser beam welding using SPH \\
    \cite{Baumann2023} & Chip jamming in deep-hole drilling \\
    \cite{Baumann2024a} & Ejector deep-hole drilling application using SPH and DEM \\
    
    \addlinespace
    \multicolumn{2}{l}{\textbf{general but domain specific information}} \\
    \cite{Monaghan1994} & Compressibility and stabilization analysis for SPH \\
    \cite{Monaghan2000} & Common stabilization schema for tensile instability in SPH \\
    \cite{Monaghan2003} & Fluid impact on rigid bodies using SPH \\
    \cite{Monaghan2005} & Summary on SPH including heat conduction treatment \\
    \cite{Liu2010} & Literature overview on SPH \\
    \cite{Violeau2016} & Overview of SPH for free-surface flows\\
    \cite{Lind2020} & Review of SPH\\
    \cite{Vacondio2020} & Challenges for SPH numerical schemas \\
  \end{tabular}
\end{table}

Information from the additional knowledge is only retrieved for some prompts (1.3, 1.4, 2.5, 2.6, 5.3, 5.4, and 5.5), whereas for the rest, the same information as before is supplied to the LLM by AnythingLLM. For the questions about appropriate particle choices or on how material properties are defined, the additional information does not bring any benefit as it contains no specifics on setting up input models but rather abstract modeling information. However, for prompt 5.3 about appropriate values for the sound velocity, a significant improvement of the LLMs reply can be seen. The common choice of choosing a value of ten times the maximum bulk fluid velocity \cite{Monaghan1994} is correctly taken from the additional literature. Furthermore, the LLMs put this choice into context with the Weakly Compressible SPH formulation when the public knowledge is supplied.

Abstract project reports or standard literature do not help to provide specific information about the software, thus do not lead to improved replies of the LLMs in these cases. However, the test shows that these additional documents can help the LLMs to improve their replies for answering general questions, especially regarding parameters of the simulation method, e.g. parameters for stabilization schemas, which are explained in the literature. For further extension of the systems, additional literature concerning a broader spectrum of SPH and DEM should be added.

\subsubsection{Adjusted Augmentation with Tailored Information}

The analysis of the retrieved information during the previous tests showed that the information provided to the LLMs is insufficient. To validate this assumption and test what the different LLMs can achieve when provided with better information, manually the required knowledge is assembled exemplarily for selected prompts. Thereby, the challenges in retrieving the necessary information and the augmented generation of the prompt responses can be separately investigated.

As test the prompts 2.4 and 3.1 are selected, which were flawed by incomplete information retrieval from the provided documents (prompt 2.4, CAD import) and incomplete information about the automatic dissection of the nested sample triangle (prompt 3.1). The manual tailored information, therefore, is build of three sections: complete sections from the documentation which are relevant for the respective prompt, the API reference for every necessary component, and corresponding minimal input examples for the components. The system prompt from AnythingLLM and the user prompt are appended to the information and the LLMs are queried.

With the tailored information provided, the results of the local LLMs are significantly improved. The smaller LLMs tended to give shorter answers compared to Gemma~3 27B. All LLMs correctly extracted the steps to import a CAD geometry in the input model from the provided document. They also deduced the automatic dissection of the triangles until they have a smaller surface area than a given limit. However, the smaller LLMs also had difficulties with the used variables names in the example, to which they attempted to establish a connection.

\subsubsection{Error-Driven Refinement}

The results of NotebookLM for prompt 6.2, to create a complete simulation model, were impressive. However, the simulation model contained several minor issues which are pointed out to the LLM for improvement. Different types of error feedback are used for this. First, a simulation component was incompletely defined, the resulting error output of the simulation run is given to NotebookLM with the assignment to fix it. Next, the use of an outdated SPH scenario and the missing of the neighborhood search are pointed out, while providing the complete component names and the desired smoothing kernel function. Finally, the LLM is instructed to remove a density pre-filtering which is unnecessary for the requested simulation.

All problems were fixed by the LLM upon request, however comments in the simulation model were either added or removed by the LLM in each round. Furthermore, an optional gradient correction schema was inserted into the simulation model by the LLM without asking for it. In conclusion, providing additional information, e.g. about issues in previous responses, can help the LLM and the RAG system to improve their response. This is described in literature as few-shot learning \cite{Brown2020}. However, especially for knowledge-intensive problems, the results need to be checked carefully to validate the changes done.

\subsubsection{Influence of the Chat History}

The previous tests are carried out in the same chat. By default, the last 20 messages are kept and sent to the LLM as chat history by AnythingLLM. The length of chat history considered in NotebookLM is not available, the very large context limit of Gemini 2.5 Flash, 1 million tokens, nevertheless, would allow NotebookLM to consider the complete chat history to a large extent. The previous tests showed that for prompt 5.4, the local LLMs often tried to create all particle separately by pure redundant definitions instead of using replicator components, although the reply to prompt 1.5 contained the correct mentioning of the necessary replicator component. However, given the size of simulation models used in the prompts 4.1 till 4.5, it is likely that the chat history is cut off due to maximum context size sent to the LLM and not due to the maximum chat history of 20 messages. To test the influence of the chat history on the result quality, and to see, if the LLMs are able to use the chat history to improve the reply quality, prompts 1.1, 1.2, 1.3, 1.5, 2.1, and 5.4 are executed in sequence in a separate chat.

On testing the shortened prompt sequence, the LLMs start using the replicator filter to create the requested cube in prompt 5.4 instead of redundant repetition of defining a particle. However, the LLMs still hallucinate on its actual syntax. On closer inspection, further issues are revealed as the information retrieved for prompt 1.5 does not contain any valuable information about the syntax and structure of the replicator component. In consequence, the LLMs are able to apply complex model definitions if the necessary informations are available.

\section{Generalizability of the Results}
\label{sec:transfer}

The work presents a focused study on the applicability of LLMs for closed-source software using the example of the mesh-free simulation software Pasimodo. The areas of application of such an LLM-based system for users as shown here can also be found in other simulation software, especially answering basic questions based on existing documentation. The other use cases shown, structured text extraction, model summarization, component explaining, and creation of simulation models, are only applicable to script-based input models that are available in text-based form. In the presented tests, the LLMs knew nothing about the Pasimodo-specific syntax of the input models, therefore, sufficient examples need to be provided to the LLMs to recognize and use the syntax. In consequence, it should be applicable to any similar simulation software.

Nevertheless, a certain amount of documentation and input examples are required. Further studies are needed to investigate whether the source code of a closed-source software can be used to derive the information necessary for creating the respective input models using LLMs. No statement can be made regarding applicability to graphics-only applications where model files are stored in binary form. Whether any graphical input can be used here must first be investigated.

For the further development of open-weight LLMs with regard to providing support for closed-source software, the largest possible context should be enabled within which connections can be easily identified. For a larger maximum input length the LLM can handle, larger inputs, thus larger and more text chunks, could be provided.

Further improvement of the user experience of the tested RAG systems would be desirable. NotebookLM and AnythingLLM are designed as chat interfaces with a linear history or order. In general, the chat history is also provided as context for later prompts. As the LLMs deliver also unintended results, the ability to revert the history or to branch of a response in multiple separate follow-up prompts would be beneficial.

Furthermore, the possibility of larger LLMs actively calling tools independently is an interesting option. Literature and our test with NotebookLM showed, that LLMs are able to self-correct generated response based on provided error messages. An option would, therefore, be to call the simulation framework itself during generation of simulation models to validate the response or to correct any upcoming error response independently. Additionally, allowing LLMs to retrieve information as needed or additional lookups for the components used in input examples offer possible improvements for the system.

Additional potential for improving the responses offers the fine-tuning the LLM by training it on new data. Especially for the specific format of Pasimodo input files, this fine-tuning could be carried out on the set of Pasimodo input examples. However, the before-mentioned challenges of the required computational effort remain.

\section{Conclusions and Outlook}
\label{sec:conclusions}

Large Language Models have become very useful tools in creating text or developing scripts and software. Information that has not been part of their training sets is not available for them. Especially for closed-source software, LLMs cannot be applied as they lack the information for solving tasks about them.

Retrieval-Augmented Generation presents an approach to overcome this issue by providing additional information to the LLM along the user prompt. The application cases of such an LLM-supported system for closed-source simulation frameworks can be found in text extraction for question answering, structured text extraction, model summarization, component explaining, and compositional reasoning to support users in creating input models for these closed-source simulation frameworks and in the creation of simulation components up to complete models.

A commercial implementation, NotebookLM, and an open-source version, AnythingLLM, of such a RAG system have been tested in this work. Whereas NotebookLM is powered by the LLM Gemini 2.5 Flash from Google, AnythingLLM can be set up using a local LLM allowing to keep all information locally. As local LLMs, the open-weight Llama~3.2 3B model from Meta, and the open-weight Gemma~3 models from Google with 3B and 27B parameters were used. The work focuses on data-protection using only information available to general users, such as students. In order to enable smaller organizations to use this technology while protecting their data, this study only uses local LLMs that can be run on consumer-grade GPUs.

The study was carried out using the closed-source simulation software Pasimodo. The RAG systems were supplied with a collection of input examples of Pasimodo, the documentation including multiple tutorials for creating models with Pasimodo, and the API reference of available components in input models. A collection of tasks covering all six areas of LLM support was created.

NotebookLM performed best from the analyzed LLMs. It was even able to distinguish particle types, detect ambiguities in the users prompts, and suggested required plugins for certain input model components. However, its limited prompt-length prevents the summarization of large models. Noticeable was the high number of sources used by NotebookLM to base its reply on. This led to impressive quality of the test results.

In contrast, AnythingLLM retrieved only four sources from its database by default, which were the same for all tested local LLMs. The small LLMs showed significant hallucinations to fill in the blanks, when documentation was incomplete or split across chunks. Especially Llama~3.2 went repeatably off-topic or confused earlier chat history with current input. The larger LLM, Gemma~3 27B, was better able to understand the retrieved information and to combine independent knowledge.

All local LLMs suffered from the insufficient information retrieved by AnythingLLM.

Different approaches to improve the response quality were tested. In particular, manually tailoring the information provided for augmented generation proved to be successful, enabling even small LLMs to compile the necessary information and answer questions more accurately. Additional knowledge in the form of project reports and standard literature can also help to answer questions related to general method orientated questions for which essential knowledge is missing in the provided documentation.

In general, the RAG systems are not immune against outdated information in the provided documents. The LLM can be guided to correct errors based on feedback from simulation outputs or based on user prompts. However, results should be checked carefully, as the LLM changes the provided simulation model in places besides the actual task.

In summary, the application of LLMs for closed-source software requires sufficient information, especially minimal input examples as otherwise syntax hallucinations occur. Additional literature helps for general aspects which are not covered in the documentation, but are method-wise well documented in standard literature. The larger models understood the provided information better and understood certain abbreviations, e.g. VTU, based on their pretraining. However, smaller models were also able to provide good answers with sufficient information provided. An adjustment of the system prompt can be beneficial to focus LLMs on the considered closed-source software, but further studies required as the tested adjustment seemed to be too restrictive.

The main challenge remains a proper information retrieval, in general, a simple RAG search with similarity does not work sufficiently. For the example of simulation software, the supplied information for generation should be drawn separately from the categories, API reference, minimal examples, documentation, and literature. Furthermore, an improvement, which does not excessively increase resource requirements, could be to retrieve besides the information chunks with the highest similarity also neighboring chunks to prevent truncated information.

The tests provided insights for the further application of LLMs and RAG systems. The tests covered potential applications in leveraging LLMs for closed-source software. In general, the problem must be clearly described regarding dimension and resolution. The results for the examples also reveal that the documentation is lacking information in some points. An example for this is the choice of the artificial sound velocity, a simulation parameter for which the relevant literature is usually referred. Therefore, future work must also include the improvement of the used information sources. An approach to supply additional non-software-specific documents showed promising results.

In summary, the out-of-the-box results for the RAG systems tested are impressive. Asking questions and generating simulation models in natural language offers immense support and time reduction for the user, especially for new users of a simulation software like Pasimodo. The tests show, however, that challenges remain. The RAG system revealed gaps in the sources provided to it, as well as in its internal source retrieval. Furthermore, the potential of fine-adjustments like adapting the retriever to provide larger text chunks or complete documents are questions of further research.

\begin{acknowledgements}
This research was supported by the Deutsche Forschungsgemeinschaft (DFG) under grant numbers 405605200 (EB 195/30-1 and EB 195/30-3).

\end{acknowledgements}

\section*{Usage of LLMs}
Apart from the experiments performed with LLMs, as clearly stated in the paper, we used LLMs
for conceptualization, spell checking, text improvement, and translation of text snippets from our mother tongue to English.

%
\section*{Conflict of interest}
The authors declare that they have no conflict of interest.

\section*{Appendix}

\begin{lstlisting}[label=lst:prompt31_code,caption={Code used in prompt 3.1 for component explaining.},language={XML},linewidth=\linewidth]
<Rectangle
  corner1 = "('4','0.0',-0.1)"
  corner2 = "('4','0.0',0.1)"
  corner3 = "('4','(ny_fluid+ny_air)*init_dx',0.1)"
  corner4 = "('4','(ny_fluid+ny_air)*init_dx',-0.1)"
>
  <Nested name = "sample_triangle">
    <Auto_Dissect_Triangle
      max_surface = "'surf'"
      color       = "(128,128,128)"
    />
  </Nested>
</Rectangle>
\end{lstlisting}

\begin{lstlisting}[label=lst:prompt32_code,caption={Code used in prompt 3.2 for component explaining.},language={XML},linewidth=\linewidth]
<VTU_Output
  enabled            = "'has_vtu_triangles'"
  use_relative_paths = "true"
  output_polyhedra   = "false"
  output_triangles   = "true"
  interval           = "'h5_dt'"
  data_dir           = "triangles"
/>
\end{lstlisting}

\begin{lstlisting}[label=lst:prompt33_code,caption={Code used in prompt 3.3 for component explaining.},language={XML},linewidth=\linewidth]
<Macro filename = "post.ma">
  <Arguments
    h5_dt              = "'h5Dt'"
    has_post_data      = "'hasPostOutput_data'"
    has_h5_SPHparticle = "'has_h5_SPHparticle'"
    has_vtu_triangles  = "'has_vtu_triangles'"
  />
</Macro>
\end{lstlisting}

\begin{lstlisting}[label=lst:prompt34_code,caption={Code used in prompt 3.4 for component explaining.},language={XML},linewidth=\linewidth]
<SPH_Integrator_PC_Leapfrog
  init_step_size      = "'init_dt'"
  max_step_size       = "'max_dt'"
  gamma               = "0.0"
  dim                 = "'dimension'"
  courant_safety      = "'courant'"
  contact_radius_mode = "'contact_radius_mode'"
  use_for_types       = "'integrator_forced_types'"
/>
\end{lstlisting}

\bibliographystyle{spmpsci}      


\end{document}